\documentclass{article}
\usepackage{spconf,amsmath,graphicx}
\usepackage{graphicx}
\usepackage{caption}
\usepackage{stfloats}
\usepackage{multirow}
\usepackage{times}
\usepackage{epsfig}
\usepackage{amsmath}
\usepackage{amssymb}
\usepackage[ruled,linesnumbered]{algorithm2e}
\usepackage{graphicx}
\usepackage{algorithmic}
\usepackage{threeparttable}
\usepackage{subfigure}
\usepackage{ifpdf}
\usepackage{cite}
\usepackage{booktabs}
\usepackage{threeparttable}
\usepackage{bbding}
\usepackage{pifont}
\usepackage{color}
\usepackage{array}
\usepackage{tabularx}
\usepackage{makecell} 

\title{Speech Swin-Transformer: Exploring a Hierarchical Transformer \\with Shifted Windows for Speech Emotion Recognition}
\name{Yong Wang$^{*,1,2}$, Cheng Lu$^{*\thanks{$^*$ The authors contributed equally to this work.},1,3}$, Hailun Lian$^{1,2}$, Yan Zhao$^{1,2}$, Bj{\"o}rn~W.~Schuller$^4$, Yuan Zong$^{\dag\thanks{$^\dag$ Corresponding Athours.},1,3,5}$, Wenming Zheng$^{\dag,1,3,5}$}

\address{
  $^1$Key Laboratory of Child Development and Learning Science of Ministry of Education, \\ Southeast University, Nanjing, China\\
  $^2$School of Information Science and Engineering, Southeast University, Nanjing, China \\
  $^3$School of Biological Science and Medical Engineering, Southeast University, Nanjing, China \\
  $^4$GLAM - the Group on Language, Audio, \& Music, Imperial College London, UK\\
  $^5$Pazhou Lab, Guangzhou, China}
\begin{document}
\ninept
\maketitle
\begin{abstract}
%
Swin-Transformer has demonstrated remarkable success in computer vision by leveraging its hierarchical feature representation based on Transformer. In speech signals, emotional information is distributed across different scales of speech features, e.\,g., word, phrase, and utterance. Drawing above inspiration, this paper presents a hierarchical speech Transformer with shifted windows to aggregate multi-scale emotion features for speech emotion recognition (SER), called Speech Swin-Transformer. Specifically, we first divide the speech spectrogram into segment-level patches in the time domain, composed of multiple frame patches. These segment-level patches are then encoded using a stack of Swin blocks, in which a local window Transformer is utilized to explore local inter-frame emotional information across frame patches of each segment patch. After that, we also design a shifted window Transformer to compensate for patch correlations near the boundaries of segment patches. Finally, we employ a patch merging operation to aggregate segment-level emotional features for hierarchical speech representation by expanding the receptive field of Transformer from frame-level to segment-level. Experimental results demonstrate that our proposed Speech Swin-Transformer outperforms the state-of-the-art methods.


\end{abstract}
\section{Introduction}

A research hotspot of affective computing is speech emotion recognition (SER), which aims to use computers to analyze and recognize emotion categories in human speech \cite{abbaschian2021deep}. The core step of SER is to extract speech features with strong emotional discrimination \cite{lu2022domain}.

Early SER work mainly combined low-level descriptor (LLD) features \cite{wani2021comprehensive}, e.\,g., fundamental frequency, Mel-frequency-Cepstral coefficients (MFCC), and frame energy, with traditional machine learning models \cite{swain2018databases}, e.\,g., Gaussian Mixture Model (GMM), Hidden Markov Model (HMM), and Support Vector Machine (SVM), to classify speech emotions. With the rise of deep learning, a series of Deep Neural Network (DNN) methods have achieved promise performance for SER  \cite{singh2022systematic}, \cite{al2023speech}, \cite{chen2023speechformer++}, e.\,g., Convolutional Neural Network (CNN), Long Short-Term Memory (LSTM), and Transformer.

Especially, Transformer \cite{waswani2017attention} have drew wide attention in SER tasks. For instance, Wang et al. \cite{wang2021novel} first used CNN and LSTM to extract temporal emotional information, and then adopted stacked Transformer layers to enhance the emotional feature extraction. Lu et al. \cite{lu2023learning} nested the frame-level Transformer in segment-level Transformer to fuse the local and global emotion features. Wang et al. \cite{wang2023timefrequency} designed Transformer to extract local emotional feature in time-frequency domain, respectively, and then aggregated them for the global emotional representations through multiple Transformers.

Current speech Transformers are expert in capturing the long-range dependencies within speech features by its Multihead Self-Attention (MSA), e.\,g., across frames or segments. They focuses on the intra-correlations of a sequence while ignoring the aggregation of inter-relationship across patches, especially the boundary information of different patches is neglected and cannot be united, which are key clues for speech emotion representation. Similarly, in computer vision, Swin Transformer \cite{liu2021swin} have proposed to address the patch feature aggregation under different-level by introducing the idea of receptive field expending in CNN into Transformer.

Inspired by Swin Transformer, we propose a novel hierarchical Transformer method for SER, which first uses the Transformer in a fixed local windows to extract emotional information, and then uses the Transformer in shifted windows to model the emotional correlation between local windows. Finally, the features are fused through a down-sampling operation. The above process constitutes a stage of our model, and we have greatly improved the emotional discrimination of features through multiple stages.

\begin{figure*}[t]
\centering
\includegraphics[width=4.4in, height=3.3in]{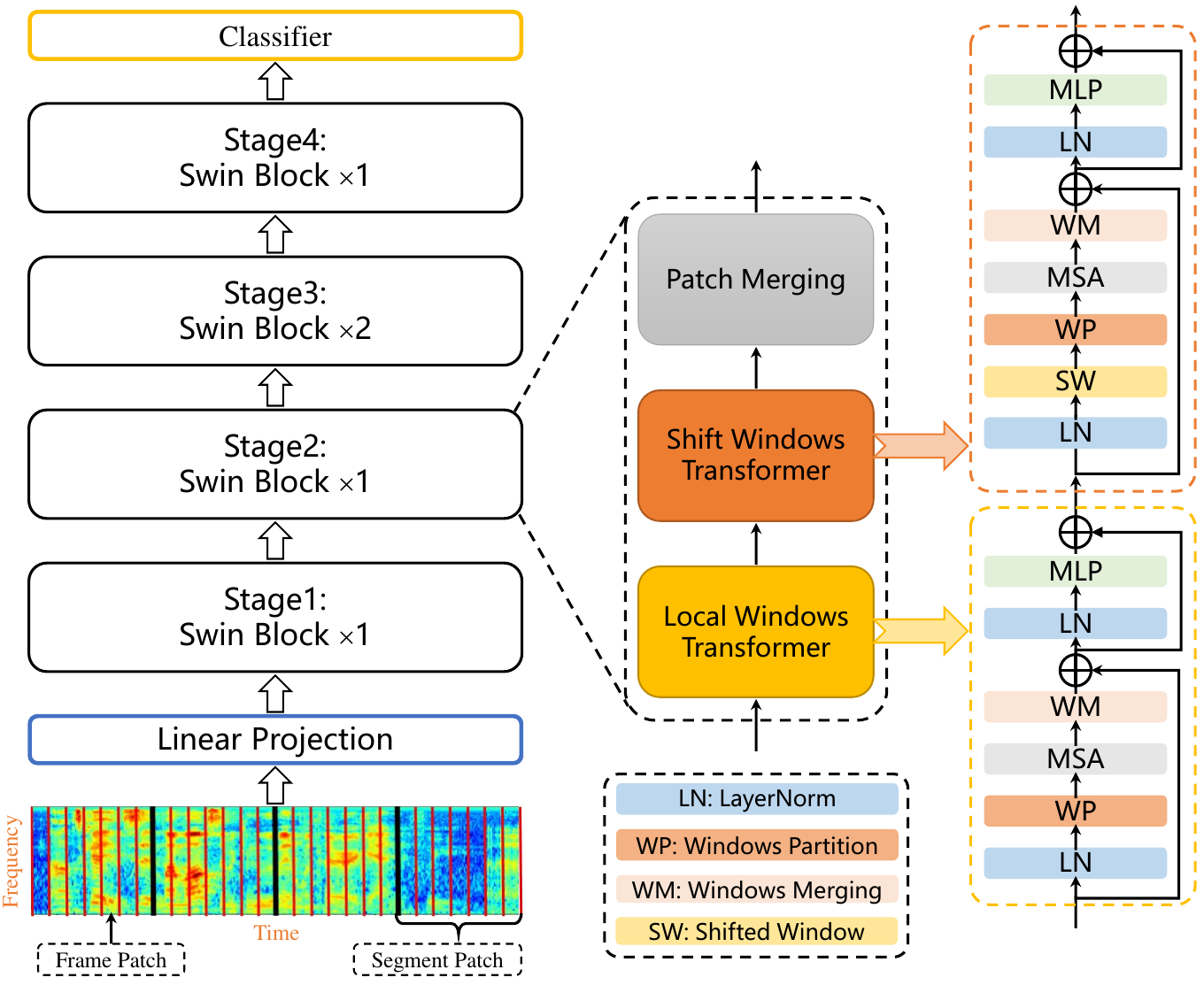}
\caption{Overall architecture of Speech Swin-Transformer for speech emotion recognition (SER).}

\label{fig:framework}
\end{figure*}

\section{Proposed Method}

Our proposed Speech Swin-Transformer is shown in Figure 1, which consists of four stages. Each stage mainly includes three modules: local windows Transformer, shifted windows Transformer, and patch merging module. This section will introduce their design details and functions.

\subsection{Local windows Transformer}

The main role of this module is to calculate the emotional correlation in each time-domain window through MSA mechanism in Transformer \cite{waswani2017attention}, thereby increasing the model's attention to emotional related fragments.

We use the log-Mel-spectrogram features $\boldsymbol{x}\in\mathbb{R}^{b \times c \times f\times d}$  extracted from the speech as the input of the model, where \emph{b} is the number of samples for each batch of model training, \emph{c} is the number of input channels, \emph{f} is the number of Mel-filter banks, and \emph{d} is the number of speech frames. Before inputting $\boldsymbol{x}$ into the first local windows Transformer, we split it in the time-domain to reduce the computational complexity of the model and obtain \emph{N} blocks log-Mel-spectrogram features $\boldsymbol{\hat{x}}\in\mathbb{R}^{b \times c \times f\times (d/N)}$. Then, we linearly map $\boldsymbol{\hat{x}}$ to get $\boldsymbol{\hat{x}}^{\prime}\in\mathbb{R}^{b \times h \times e}$ as the input of the first Transformer, where $ \emph{h}=\emph{f}\times \emph{d}/\emph{N}$ and \emph{e} is the dimension of embedding.

The main difference between the local windows Transformer and the traditional Transformer is to use a window of size $[\emph{f}, \emph{t}]$ to divide the input features $\boldsymbol{\hat{x}}^{\prime}$, get \emph{M} time-domain blocks, and then calculate MSA in each window, i.\,e., $\emph{M}=\emph{d} / (\emph{t} \times \emph{N})$, \emph{f} and \emph{t} are the frequency-domain and time-domain length of the window, instead of a traditional Transformer
that directly performs MSA on the input features. After calculating the MSA,
we concatenate the windows in the original order into feature $\boldsymbol{s}\in\mathbb{R}^{b \times h \times e}$ with the same dimension as the input feature $\boldsymbol{\hat{x}}^{\prime}$. The process is as follows:
\begin{equation}
\boldsymbol{l} =  WP(LN(\boldsymbol{\hat{x}}^{\prime})),
\end{equation}
\begin{equation}
\boldsymbol{m} =  WM(MSA(\boldsymbol{l}))+\boldsymbol{\hat{x}}^{\prime},
\end{equation}
\begin{equation}
\boldsymbol{s} =  MLP(LN(\boldsymbol{m}))+\boldsymbol{m},
\end{equation}
where $LN(\cdot)$ and $WP(\cdot)$ are layer normalization and window partition, respectively. The $MSA(\cdot)$, $WM(\cdot)$, and $MLP(\cdot)$ represent MSA, window merging, and the fully connected layer operation, respectively. Besides, $\boldsymbol{l}$ and $\boldsymbol{m}$ are the outputs of the window partitioning and window merging modules within the Transformer, respectively. The obtained $\boldsymbol{s}$ serves as the input for the shifted windows Transformer.

\subsection{Shifted windows Transformer}

Since the local windows Transformer module only performs MSA in each window, the connection between windows is ignored. We design a Transformer with moving windows to focus on cross-window connections. The main difference between this module and the local windows Transformer is that there is an additional shift operation before the window partition, and the moving step is set to half length of the window's time-domain. The process can be represented as
\begin{equation}
\boldsymbol{\hat{l}} =  WP(SF(LN(\boldsymbol{s}))),
\end{equation}
\begin{equation}
\boldsymbol{\hat{m}} =  WM(MSA(\boldsymbol{\hat{l}}))+\boldsymbol{s},
\end{equation}
\begin{equation}
\boldsymbol{\hat{s}} =  MLP(LN(\boldsymbol{\hat{m}}))+\boldsymbol{\hat{m}},
\end{equation}
where $SF(\cdot)$ represents window movement similar to Swin Transformer \cite{liu2021swin}. The $\boldsymbol{\hat{l}}$ and $\boldsymbol{\hat{m}}$ are the outputs of the window partitioning and merging modules within Transformer, respectively. The final output $\boldsymbol{\hat{s}}\in\mathbb{R}^{b \times h \times e}$ of Transformer module will be used as input for the patch merging module.

\subsection{Patch Merging Module}

This module reduces the resolution of features and adjusts the number of channels through downsampling operations, which can reduce computational complexity and form a hierarchical structure \cite{liu2021swin}. Similar to Swin transformer, the specific approach is to select elements at intervals of one point in the time-domain and frequency-domain of the features, and then concatenate them together as a whole tensor. The channel dimension will become four times the original size, and then be adjusted to twice the original size through a fully connected layer. The operations can be represented as
\begin{equation}
\boldsymbol{\hat{s}}^{\prime} = PM(\boldsymbol{\hat{s}}),
\end{equation}
where $PM(\cdot)$ represents patch merging. The output feature $\boldsymbol{\hat{s}}^{\prime}\in\mathbb{R}^{b \times (h/4) \times 2e}$ will be used as the input for the local windows Transformer in the next stage.

The output of the shift windows Transformer of the last stage will be normalized, average pooled, and flattened to obtain feature $\boldsymbol{y}\in\mathbb{R}^{b \times 8e}$, without going through the patch merging module. We pass the obtained feature $\boldsymbol{y}$ through the fully connected layer to calculate the predicted emotional probability; the operation is as follows:
\begin{equation}
\boldsymbol{\hat{y}} =  FC(\boldsymbol{y}),
\end{equation}
where $FC(\cdot)$ represents a fully connected layer. We use the Softmax function \cite{dubey2022activation} to make the emotion with the highest prediction probability among the fully connected layer output features $\boldsymbol{\hat{y}}\in\mathbb{R}^{b \times k}$ as the predicted emotion. Then, we optimize the model by reducing the cross-entropy loss \emph{Loss}  between predicted emotion labels and true emotion labels $\boldsymbol{z}\in\mathbb{R}^{b \times k}$, where \emph{k} is the number of categories of emotions. We can represent the process as
\begin{equation}
\boldsymbol{\hat{y}}^{\prime} =  Softamx(\boldsymbol{\hat{y}}),
\end{equation}
\begin{equation}
Loss = CrossEntropyLoss(\boldsymbol{\hat{y}}^{\prime},\boldsymbol{z}),
\end{equation}
where $Softmax(\cdot)$ and $CrossEntropyLoss(\cdot)$ are the Softamx function and Cross-Entropy loss function \cite{lu2022neural}, respectively.

\section{Experiments}

\subsection{Experimental Databases}

Two widely used databases, i.\,e., IEMOCAP \cite{busso2008iemocap} and CASIA \cite{zhang2008design} are adopted to verify the effectiveness of the proposed method. IEMOCAP is a multimodal dataset that includes data such as video, voice, and text transcription, released by the SAIL Laboratory of the University of Southern California. It contains multimodal data of 5 male actors and 5 actresses in the process of emotional binary interaction. Our experiments select 4 types of emotional samples (\emph{angry}, \emph{happy}, \emph{neutral}, \emph{sad}) in impromptu scenes and a total of 2280 audio data.
CASIA is a Chinese emotional speech database established by the Institute of Automation, Chinese Academy of Sciences. Under the fixed script content, 4 volunteers (2 men and 2 women) perform 6 emotions (\emph{angry}, \emph{happy}, \emph{fear}, \emph{sad}, \emph{neutral}, and \emph{surprise}) in this database, and the sampling rate of the samples is 16\,000\,Hz. In this paper, 200 samples of each emotion are selected, and a total of 1200 samples of audio data are used for our experiments.

\subsection{Experimental Protocol}

We utilize the same Leave-One-Speaker-Out (LOSO) cross-validation protocol as previous studies \cite{lu2023learning}, \cite{lu2022speech}, \cite{wang2023timefrequency} to evaluate model performance. In CASIA, one speaker's sample is used for testing, and the remaining three speaker's samples are used for training. For IEMOCAP, the samples of one speaker are used as the test set, and the samples of other 9 speakers are used as the training set. Moreover, the weighted average recall (WAR) and unweighted average recall (UAR) \cite{stuhlsatz2011deep} are recorded as the evaluation metrics for the comparison experiments.

\subsection{Experimental Setup}

In order to expand the number of samples and maintain the integrity of speech emotions, we sliced the audio samples into 128 frames (20\,ms per frame) of fragments. Then, we pre-emphasize speech segments with a coefficient of 0.97. We use Short-Time Fourier Transform (STFT) to extract the log-Mel-spectrogram of speech, which uses a 20\,ms Hamming window with a frame shift of 10\,ms. Besides, the number of points of Fast Fourier Transform (FFT) and bands of Mel-filter are 512 and 32,  respectively. Finally, we obtain the feature
dimension
of \emph{b}=64, \emph{c}=1, \emph{f}=32, and \emph{d}=128 as the input features for the proposed model.

The model parameters in this paper are set as follows: \emph{N}=4, \emph{t}=4, \emph{e}=96. The number of Transformer in the 4 stages of our model is 2, 2, 4, 2 respectively. The proposed model is implemented by the Pytorch framework \cite{paszke2019pytorch} winth NVIDIA A10 GPUs. And its parameters are updating by a Adam \cite{kinga2015method} optimizer with a learning rate of 0.0001 and the training epoch is set to 100.

\subsection{Results and Analysis}

\begin{table}[t]
\caption{Comparison results on IEMOCAP and CASIA with the experimental protocol of Leave-One-Subject-Out (LOSO).}
\label{tab:experimental-results}
\begin{tabular}{|c|cc|cc|}
\hline
\multirow{2}{*}{Comparison Method} & \multicolumn{2}{c|}{\begin{tabular}[c]{@{}c@{}}IEMOCAP\\ (10 speakers) (\%)\end{tabular}} & \multicolumn{2}{c|}{\begin{tabular}[c]{@{}c@{}}CASIA\\ (4 speakers) (\%)\end{tabular}} \\ \cline{2-5}
                                   & \multicolumn{1}{c|}{WAR}   & UAR      & \multicolumn{1}{c|}{WAR}    & UAR    \\ \hline \hline
ViT                                 & \multicolumn{1}{c|}{70.22}    &58.58     & \multicolumn{1}{c|}{42.83 }    &42.83     \\ \hline
TNT                                  & \multicolumn{1}{c|}{70.61}    &59.72     & \multicolumn{1}{c|}{46.58}    &46.58     \\ \hline
Swin-Transformer\ \cite{liu2021swin}        & \multicolumn{1}{c|}{70.75}    &60.08     & \multicolumn{1}{c|}{47.25 }    &47.25     \\ \hline
LGFA \cite{lu2023learning}                & \multicolumn{1}{c|}{73.29}    &62.63     & \multicolumn{1}{c|}{49.75 }    &49.75     \\ \hline
ATFNN \cite{lu2022speech}                 & \multicolumn{1}{c|}{73.81}    &64.48     & \multicolumn{1}{c|}{48.75}    &48.75     \\ \hline
TF-Transformer\ \cite{wang2023timefrequency}& \multicolumn{1}{c|}{74.43}    &62.90     & \multicolumn{1}{c|}{53.17}    &53.17     \\ \hline \hline
Speech Swin-Trans. (Ours)              & \multicolumn{1}{c|}{\textbf{75.22}}  & \textbf{65.94}  & \multicolumn{1}{c|}{\textbf{54.33}}  &\textbf{54.33}  \\ \hline
\end{tabular}
\end{table}

\begin{figure}[htb]
\centering
\subfigure[Confusion matrix on IEMOCAP]{
\label{Mel_spec_iemocap}
\includegraphics[width=0.485\linewidth]{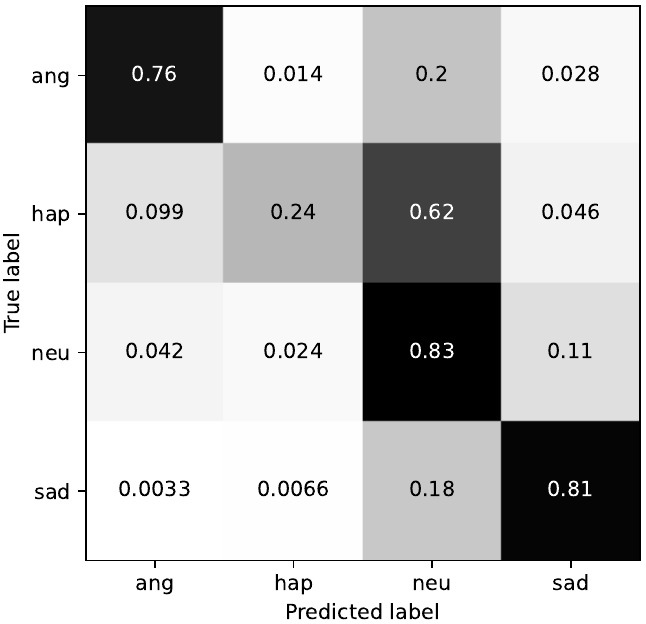}}
\subfigure[Confusion matrix on CASIA]{
\label{T_Trans_Attention_iemocap}
\includegraphics[width=0.485\linewidth]{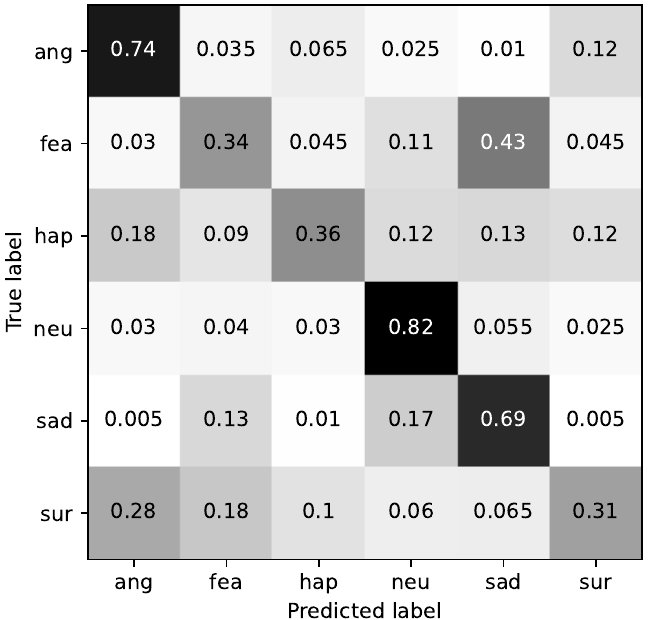}}
\caption{Confusion matrices of Speech Swin-Transformer. }
\label{fig:cm}
\end{figure}

To compare the performance of our proposed Speech Swin-Transformer, we selected some state-of-the-art algorithms based on Transformer architecture for performance comparison, i.\,e., Vision Transformer (ViT) \cite{dosovitskiy2021an}, Transformer in Transformer (TNT) \cite{han2021transformer}, a hierarchical vision Transformer using shifted windows (Swin-Transformer) \cite{liu2021swin}, local to global feature aggregation learning based on Transformers (LGFA) \cite{lu2023learning}, as well as SER via an attentive time–frequency neural network (ATFNN) \cite{lu2022speech}, and a novel time frequency joint learning method (TF-Transformer) \cite{wang2023timefrequency}. Note that ViT and TNT are implemented by ourselves.

The experiments on IEMOCAP and CASIA are shown in Table~\ref{tab:experimental-results}. We can observe that our proposed method achieves the state-of-the-art results. In detail, compared with other methods under the Leave-One-Speaker-Out protocol, our Speech Swin-Transformer achieves the highest WAR (75.22\%) and UAR (65.94\%) on the IEMOCAP dataset. Furthermore, we also calculate the confusion matrix obtained on IEMOCAP for comprehensive comparison, shown in Fig.~\ref{fig:cm}(a). Through the confusion matrix on IEMOCAP, it is obvious that our model has a high recognition rate (over 75\%) in most emotions (\emph{angry}, \emph{neutral} and \emph{sad}), which verifies the effectiveness of the proposed Speech Swin-Transformer.

%

Moreover, we can also find that the recognition rate of our model on \emph{happy} is only 24\%, and it is easily misclassified as \emph{neutral}. The possible reason is that the number of samples in the IEMOCAP dataset is unbalanced. The number of \emph{happy} samples is only 284, which is the smallest number of samples in IEMOCAP. This makes the model unable to learn the unique emotional characteristics of \emph{happy}, making \emph{happy} easy to be misclassified for \emph{neutral}. In addition, it is believed that \emph{happy} and \emph{neutral} are both high valence emotions \cite{yang2022hybrid}. They are similar in terms of valence dimensions, which may lead to misclassification.


According to the experimental results in Table~\ref{tab:experimental-results}, it is obvious that our method also achieves the best WAR (54.33\%) and UAR (54.33\%) on CASIA. Based on the experimental results, we can discover some interesting things. Firstly, the UAR and WAR of the CASIA dataset experimental results are the same because the sample size for each type of emotion is balanced. In addition, we can see that compared with other advanced methods, the performance of our model improves by at least 1.16\% on WAR and UAR.
%
%

\begin{figure*}[t]
\centering
\includegraphics[width=0.75\linewidth]{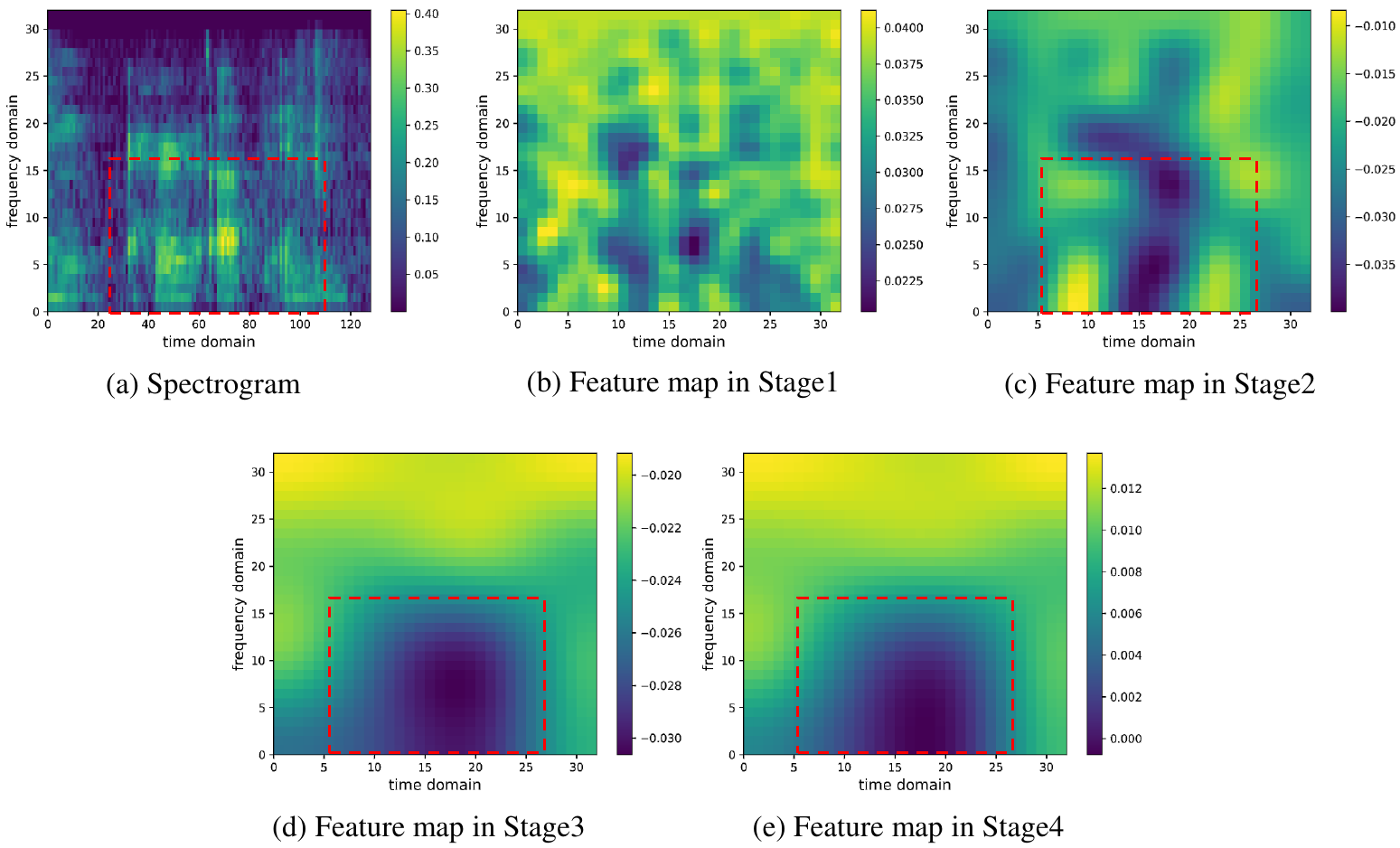}
\caption{Visualizations of hierarchical feature maps generated by the different stages of Speech Swin-Transformer.}
\label{fig:framework}
\end{figure*}

Fig.~\ref{fig:cm}(b) shows the confusion matrix of our Speech Swin-Transformer on CASIA, in which we can find that the recognition accuracy of our model on \emph{angry}, \emph{neutral}, and \emph{sad} in CASIA exceeds 68\%, which reflects the effectiveness of the proposed method.

Note that the recognition rates of our proposed method for the three types of emotions (\emph{fear}, \emph{happy} and \emph{surprise}) are all lower than 40\%. Specifically, \emph{fear} is easily misidentified as \emph{sad}, which may be due to the fact that these two types of emotions are low valence emotions \cite{yang2022hybrid}, and similar valences may lead to this experimental phenomenon. In addition, \emph{happy} and \emph{surprise} are both prone to being misclassified as \emph{angry}, as these three types of emotions are high arousal emotions \cite{yang2022hybrid}, and similar arousal levels may lead to misidentification of emotions.


\subsection{Hierarchical Feature map visualizations}
To further investigate the hierarchical representation results of the Speech Swin-Transformer, we also visualized the feature mappings in the four stages, as shown in Figure 3. From (a) to (e), we can observe that the network's shallow layers (Stage 1) focus on the general features of the spectrogram, e.\,g., color and shape. As the stages progress, the network gradually differentiates the parts of the spectrogram that contain semantic information from those that do not (as shown by the red box in (c)). In the deeper Stage 3 and 4, the network distinguishes the responses in the mid-low frequency range, as indicated by the red box, which are closely related to the expression of \emph{sad}. This phenomenon is consistent with previous studies \cite{lu2022speech}, that is low arousal emotions (e.\,g., \emph{sad}) commonly exhibit activations in the mid-low frequency range, demonstrating that our method can indeed obtain hierarchical speech emotion representations.

\section{Conclusion}

In this paper, we proposed a Speech Swin-Transformer for SER based on the hierarchical Transformer architecture, which can effectively aggravate the emotional features at different-levels. Our experiments on IEMOCAP and CASIA demonstrate the effectiveness of our method. Compared with other state-of-the-art architectures based on the Transformer model, our model has improved to varying degrees in both WAR and UAR. Future research will focus on the patch boundary modeling both on frequency domain and time-frequency domain.

\section{Acknowledgements}

This work was supported in part by the National Key R\&D Project under the Grant 2022YFC2405600, in part by the NSFC under the Grant U2003207 and 61921004, in part by the Jiangsu Frontier Technology Basic Research Project under the Grant BK20192004, in part by the YESS Program by CAST (Grant No. 2023QNRC001) and JSAST (Grant No. JSTJ-2023-XH033), in part by the ASFC under the Grant 2023Z071069003, in part by the China Postdoctoral Science Foundation under the Grant 2023M740600, and in part by the Jiangsu Province Excellent Postdoctoral Program.



\small
\bibliographystyle{IEEEbib}
\bibliography{reference}

\end{document}